\documentclass{article}

\usepackage{PRIMEarxiv}

\usepackage[utf8]{inputenc} % allow utf-8 input
\usepackage[T1]{fontenc}    % use 8-bit T1 fonts
\usepackage{hyperref}       % hyperlinks
\usepackage{url}            % simple URL typesetting
\usepackage{booktabs}       % professional-quality tables
\usepackage{amsfonts}       % blackboard math symbols
\usepackage{nicefrac}       % compact symbols for 1/2, etc.
\usepackage{microtype}      % microtypography
\usepackage{lipsum}
\usepackage{fancyhdr}       % header
\usepackage{graphicx}       % graphics
\graphicspath{{media/}}     % organize your images and other figures under media/ folder
\usepackage{amsmath,amssymb,amsfonts}
\usepackage{algorithmic}
\usepackage{graphicx}
\usepackage{algorithm,algorithmic}
\usepackage{hyperref}
\hypersetup{hidelinks=true}
\usepackage{subcaption}
\usepackage{multirow} 
\title{Impact of Age Specialized Models for Hypoglycemia Classification
}

\author{
  Beyza Cinar\\
  Data Engineering\\
  Helmut Schmidt University\\
  Germany\\
  \texttt{cinarb@hsu-hh.de} \\
   \And
  Maria Maleshkova\\
  Data Engineering\\
  Helmut Schmidt University\\
  Germany\\
  \texttt{maleshkm@hsu-hh.de} \\\\
}

\begin{document}
\maketitle

\begin{abstract}
Disease progression varies with age and is influenced by underlying genetic, biochemical, and hormonal etiologies, suggesting the need for tailored monitoring, care, and medication beyond standard clinical guidelines. Specifically, in autoimmune diseases like type 1 diabetes (T1D), where patients depend on exogenous insulin to compensate for insulin deficiency, medication dosing and the physiological response reflected in vital signs can differ. Insulin therapy can lead to hypoglycemia, a dangerous condition characterized by decreased blood glucose levels ($\leq$70). This risk can be mitigated through improved diabetes management supported by data analytics. Notably, leveraging data from continuous glucose monitoring (CGM) devices, hypoglycemia onset can be predicted. However, while glucose variability, auto-antibody levels, and hypoglycemia occurrence differ across age groups, hypoglycemia classification most often only relies on population-based models specialized in specific age ranges. In this work, we classify hypoglycemia 0, 5-15, 20-45, and 50-120 minutes before onset using DiaData, a large CGM dataset of patients with T1D ranging from children to seniors. In particular, we investigate: 1) the generalizability of a population-based model including all age groups, 2) the impact of age-segmented models trained separately per age group, and 3) the effect of model individualization through transfer learning. The results show that a global population-based model yields similar or superior performance compared to age-segmented models. These findings suggest that data from children, teenagers, and adults can be combined for training models on hypoglycemia classification. While glucose variation differs across age groups, short-term hypoglycemic patterns are similar. However, data of children obtain their best recall with age specialized model.
\end{abstract}


\section{Introduction}
Effective alert systems warning of hypoglycemic glucose levels are essential to prevent life-threatening complications such as dizziness, loss of consciousness, coma, or even death. Hypoglycemia, characterized by glucose concentrations below 70 mg/dL, is a common side effect of insulin therapy and occurs most frequently in individuals with Type 1 diabetes (T1D) \cite{Allen2001-nd}. T1D is an endocrine disease leading to insulin insufficiency due to an autoimmune destruction of pancreatic beta cells. Affected patients depend on external insulin injections~\cite{Leete2018-nu, Parviainen2022-il}, as insulin is essential for maintaining healthy blood glucose levels \cite{elsayed_2_2023}. Hypoglycemia remains one of the leading causes of mortality among patients with T1D because the onset is usually rapid and asymptomatic, underscoring the need for timely detection strategies \cite{Allen2001-nd}.  

Diabetes management can be enhanced and personalized with the integration of continuous glucose monitoring (CGM) systems \cite{jessica_lucier_type_2023}. Data analysis and artificial intelligence (AI) based methods can further increase the utility of CGM devices by providing trend analysis, forecasting future values, or predicting anomalies like hypoglycemia \cite{Felizardo2021review, Grensing2025EarlyWarning}. Accordingly, hypoglycemia prediction has become a major research focus, intending to enable timely interventions. While most hypoglycemia classification systems, based on binary classifiers, report increased accuracy of 80\% \cite{dave_feature-based_2021, Hni2025, Felizardo2021review}, the precision often remains lower \cite{Felizardo2021review, Hni2025}. One of the main limitations is the lower proportion of hypoglycemic values relative to normal glucose values (70-180 mg/dL), resulting in data imbalance. Moreover, glucose values can be influenced by several factors, including the insulin dose, the meal intake, and the physiological and psychological states of the subject, whereas contextual data is often missing in publicly available datasets \cite{ghimire_deep_2024}. Notably, studies have demonstrated that vital signs exhibit subject-specific patterns, underscoring the importance of personalized treatments and tailored monitoring beyond standard clinical guidelines. Yang et al. illustrate that a different feature set is required per subject for maximizing glucose forecasting performance, indicating that data patterns and lifestyle are subject-dependent \cite{yang_personalized_2024}, and Bent et al. reveal that individual thresholds outperform a global threshold~\cite{bent_engineering_2021}.

Specifically, several studies report that T1D pathogenesis, epidemiology, risk, and disease progression are impacted by the patient's age \cite{Leete2018-nu, So2022-ze}, with age-related variation in the rate of auto-antibodies, and decline in C-peptide \cite{So2022-ze, Leete2018-nu}. Patients diagnosed at a younger age would experience greater islet autoimmunity from immunological, metabolic, and genetic factors, and children would experience decreased immune regulatory function \cite{Leete2018-nu}. In contrast, although depending on insulin, patients with an older age at diagnosis could retain significant numbers of insulin-containing islets \cite{Leete2018-nu}. Furthermore, significant age-related differences are reported in particular in patients below 7 and above 12 years. The first group ($<$7 years) showed higher levels of insulin autoantibodies. The third group ($>$12 years) had more severe metabolic decompensation \cite{Parviainen2022-il}. Investigating glucose variation in puberty and youth, Zhu et al. inform that pre-pubertal youth experienced higher glucose variability and increased mean glucose levels. They report that maintaining target glycemic levels remained a challenge in young persons with T1D, particularly during adolescence due to physiologic factors related to hormonal changes during puberty \cite{Zhu2020-hd}.

Moreover, among sociodemographic variables, age is reported to be the only feature that increases hypoglycemia risk. Allen et al. analyzed demographic and self-management factors via longitudinal data of blood samples and questionnaires from a cohort of children, adolescents, and young adults with T1D. Findings show that the association of frequent hypoglycemia with intensive insulin therapy increased with age. Also, better glycemic control and older age were associated with severe hypoglycemia. Approximately after the age of 15, the risk of severe hypoglycemia slowed down \cite{Allen2001-nd}. Consequently, age emerges as an important determinant, since it is related to disease progression and hypoglycemia occurrence. Therefore, incorporating age into model development may refine performance \cite{So2022-ze}.
 
In this work, we investigate the impact of age on hypoglycemia classification, building on our previous study, which demonstrated that age groups have the highest impact among personal features, including sex, age group, race, or BMI \cite{bibe_diadata_2025}. To train the models, DiaData, a large CGM dataset including 2499 subjects ranging from children (2 years) to seniors (87 years), is leveraged. In particular, we make the following contributions: 1) We investigate bins for the most distinctive age groups and assess their glucose variation. 2) We select the most distinct classes for classifying hypoglycemia up to 120 minutes before onset, ranging between 0, 5-25, 20-60, 30-120, 60-120 minutes for classes 0, 1, 2, 3, and 4, respectively. 3) We compare global population-based and age-segmented population-based hypoglycemia classification models using a Fully Convolutional Network (FCN) model. Specifically, we assess the generalizability of both approaches, testing if integrating children and adult data is feasible and whether age-group specialized models perform more efficiently. 4) Finally, we compare population-based models and, via transfer learning, individualized models. 

The remainder of this work is structured as follows: Section \ref{sec:stota} explores studies leveraging age-related data for glucose or hypoglycemia prediction, section \ref{sec:methods} depicts the dataset, age group definitions, the data preprocessing pipeline to prepare data for the classification task, and the architecture of the models. Section \ref{sec:results} shows the results for the ablation studies and the ML models. Finally, section \ref{sec:discussion} discusses the findings, and section \ref{sec:conclusion} summarizes the study.

\section{Related Work}
\label{sec:stota}

Previously, various works explored age-based models for forecasting glucose values and demonstrated superior performance. Ghimire et al. examine model generalizability across four datasets, where the glucose dynamics differ, containing children aged 6–13 years (DCLP5), subjects between 14–71 years (DCLP3 ), and adults aged 20–80 years (OhioT1DM). The RT dataset comprises all age categories. Models are trained independently on each dataset and evaluated on the others using CGM sequences over the last 2 hours to predict 30 and 60 minutes of CGM values. Results show similar performance between the train and test datasets. The DCLP5 dataset yields the best cross-dataset performance, suggesting robust generalization capability within younger populations. The RT dataset generalizes effectively except when predicting on the DCLP5 dataset. Conversely, the smaller OhioT1DM dataset performs poorly, and the DCLP3 dataset can only classify the OhioT1DM dataset well. Overall, supported by statistical analysis, the study highlights that cross-dataset training is feasible across age groups~\cite{ghimire_deep_2024}. Similarly, Ryser et al. forecast glucose levels to investigate model generalizability, but they propose a transfer learning method for the individualization of specific age ranges. They utilize the OhioT1DM, CITY (14-24 years), and DiaCamp (7-10 years) datasets. The model is pretrained with the OhioT1DM and the CITY datasets, and fine-tuned with the DiaCamp dataset. They forecast 30, 60, and 120 minutes of CGM values and demonstrate that pretraining on the OhioT1DM dataset leads to superior performance, whereas the CITY dataset does not impact performance significantly. Thus, CGM data of children from small datasets can be better predicted when being pretrained on larger~\cite{Ryser2025}. Yang et al. propose a personalized framework while categorizing patients based on sex and age. Likewise, the OhioT1DM dataset is leveraged, and glucose values of 30 and 60 minutes are predicted. They compare training a population-based model, training individual models for all subjects, and training on clusters of subjects. While personalized models are superior, they suggest cluster-based models to reduce cost~\cite{yang_personalized_2024}.

However, the impact of age levels and model generalization across age groups was only investigated for regression-based methods. It remains a research gap for hypoglycemia classification, with most approaches presenting population-based models. Dave et al. classify hypoglycemia onset using binary classification separately for various PHs ranging from 15 to 60 minutes \cite{dave_feature-based_2021}. Likewise, Hüni et al. leverage binary classification models. Their models yield low precision, indicating high false alarm rates~\cite{Hni2025}. D'Antoni et al. propose a two-step classification framework by first classifying between euglycemia, hypoglycemia, and hyperglycemia. Second, they classify the event onset 15 to 120 minutes before. Class imbalance led to worse recall, and relevant performance is only reported for PHs of 15- and 30-minute PHs \cite{dantoni_prediction_2022}. Hypoglycemia can be classified up to 15 minutes before the event with a recall of 70-80\%, but with lower precision. PHs above 30 minutes do not show reliable performance \cite{dantoni_prediction_2022, Hni2025}.

Finally, Cinar et al. report a multi-class hypoglycemia onset classification framework, integrating various PHs within a single model. Likewise, a decline in performance with increasing PHs is observed, particularly for longer horizons of 30-60 minutes \cite{cinarMaster, Cinar_BHI2025, bibe_diadata_2025}. The best performance for a 5-25 minute event onset is achieved by an FCN model trained on CGM values of DiaData \cite{Cinar_BHI2025}. Based on SHAP-based feature importance analysis, it is further demonstrated that among demographic and personal features, only the age group of subjects with T1D shows an impact on hypoglycemia classification~\cite{bibe_diadata_2025}.

\section{Methods}
\label{sec:methods}

This study explores the generalizability of global population-based models trained with all age categories as well as age-specific models. We investigate whether models tailored to age groups increase the classification performance of hypoglycemia. Leveraging the DiaData dataset, participants are segmented into distinct age groups for age-specific classification models. Subsequently, we examine whether individualized model tuning enhances prediction performance compared to population-based models. Each model is evaluated on subjects with the largest amount to ensure robust and reliable performance. In the following section, the utilized dataset, age group splits, model architecture, and validation methods are described.\footnote{The code is available at: \href{https://github.com/Beyza-Cinar/Age-Population-Generalization-Hypoglycemia}{https://github.com/Beyza-Cinar/Age-Population-Generalization-Hypoglycemia}}

\subsection{DiaData}

The models were trained on Subdatabase I of DiaData \cite{DiaData_Zenodo}, a large dataset of 2510 subjects with T1D comprising CGM data collected in 5-minute intervals, and demographics of sex and age. Participants range from children to seniors (2 to 87 years) \cite{cinar2025diadata}. The Subdataset I was curated by integrating 13 publicly available datasets collected from subjects under free-living conditions \cite{RodriguezLeon2023,DiatrendPaper,HupaMendeley,ShanghaiFigshare,ictinnovaties_diabetes_heart_rate_2025,jab_center_diabetes}. DiaData has an equal sex distribution (54\% females), and on average, each subject has data collected for 146 days (d), with a minimum of 1 day and a maximum of 1555 days. The dataset reports a substantial amount of missing CGM values, with 132 million datapoints.

\subsection{Definition of Age Groups}

\begin{figure}[!t] 
\centering
    \includegraphics[scale=0.35]{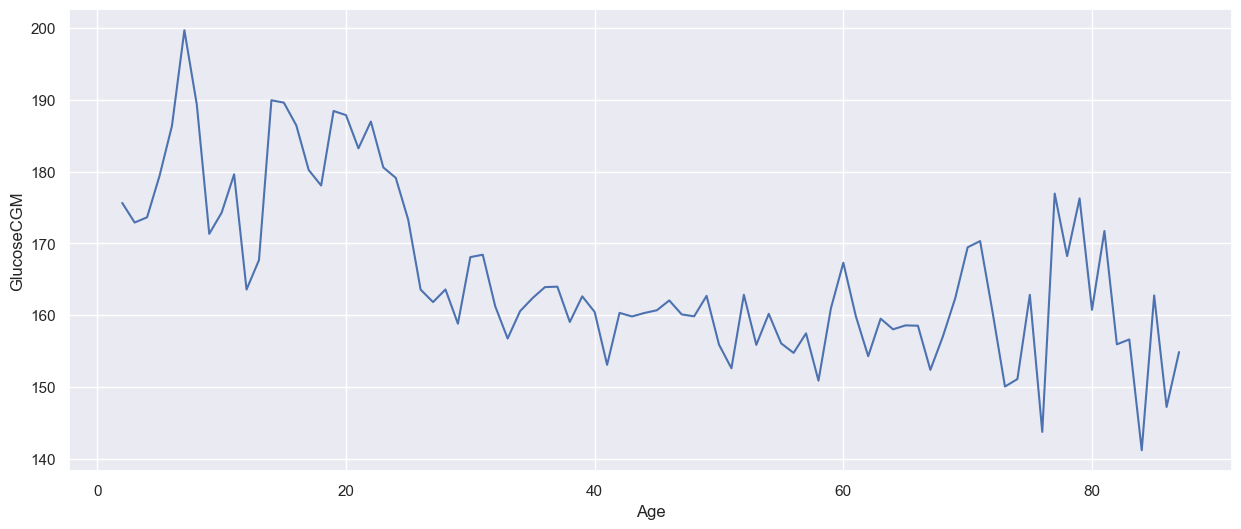}
\caption{Mean CGM per Age}
\label{fig:cgmmean}
\end{figure}

\begin{figure}[!t]
\centering
    \includegraphics[scale=0.35]{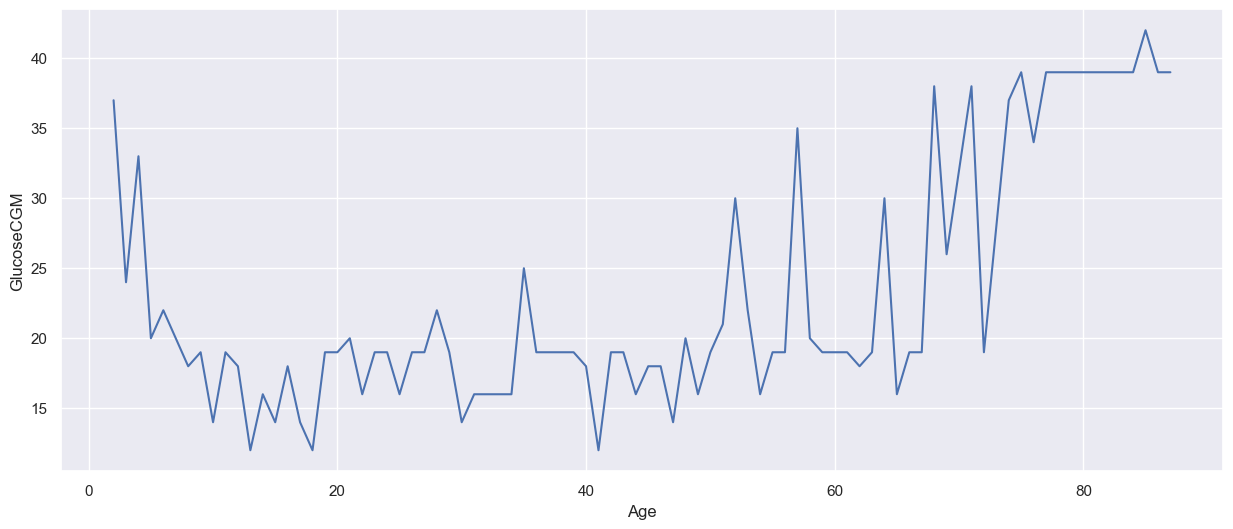}
\caption{Minimum CGM per Age}
\label{fig:cgmmin}
\end{figure}

\begin{figure}[!t]
\centering
    \includegraphics[scale=0.35]{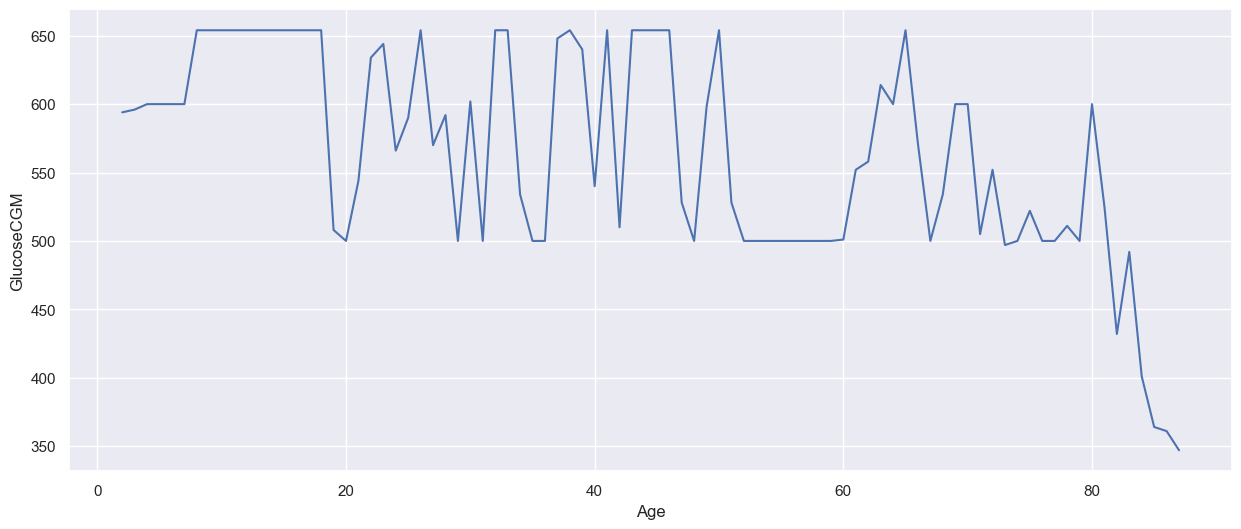}
\caption{Maximum CGM per Age}
\label{fig:cgmmax}
\end{figure}

The reported ages of patients ranged from 2 to 87 years, whereas some subjects only provided age intervals rather than specific values. For instance, the SHD dataset reports that subjects were not younger than 60 years, but the age of the oldest person is not mentioned. Consequently, for defining age groups, only numerically specified ages were considered. Age groups were defined based on two strategies. First, CGM variability was assessed visually as presented in Figures \ref{fig:cgmmean}, \ref{fig:cgmmin}, \ref{fig:cgmmax}, and then Tukey's analysis was conducted experimentally for several splits. Comparing the boxplots of each age separately, we observe similar variation and means between the ages of 2-5, 6-8, 9-13, 14-18, 19-25, 26-50, 50-73, and older than 73 years, with fewer differences between adults and seniors. Fig. \ref{fig:cgmmean}, illustrating the mean CGM value per age as a line graph, reveals increased mean values in subjects below 24 years, with a special emphasis on 6-7 and 13-20 years. Contrariwise, adults experience less variation and decreased mean values. Hence, glucose dynamics differ across age, suggesting that aggregating all subjects into a single model may not be optimal. This concern is further supported by the unequal age distribution of the dataset. In particular, prior research has shown that children and adolescents are significantly underrepresented \cite{cinar2025diadata}. The minima, demonstrated in Figure \ref{fig:cgmmin}, reveal very irregular behavior, while children, teenagers, and adults have very low glucose patterns, older adults show the highest peaks, and their glucose values do not fall below 10 mg/dL. Likewise, the maxima shown in Figure \ref{fig:cgmmax} do not reveal significant patterns and differences across the age levels. 

In addition, Tukey's analysis was conducted for different age ranges \cite{abdi2010newman}. The analysis was run on different combinations of age groups ranging from 0-13 years for children, 20-55 for adults, and 35-100 for the elderly. Also, smaller splits differentiating between children 0-10, teenagers 10-20, adults 20-30, older adults 40-55, and the elderly above 55 were tested. In all conducted experiments except the groups of 10–15 and 16–25, the null hypothesis was rejected, indicating significantly different mean CGM values across demography. The results of the best split can be seen in Table \ref{tab:Agesplit}, defining age groups of 0-13, 14-20, 21-44, 45-100. 
\begin{table}[!b]
\centering
\caption[Result of Tukey's Analysis for Best Splits]{Result of Tukey's Analysis for Best Splits\label{tab:Agesplit}}
\begin{tabular}{|c|c|c|c|c|c|}
 \hline
\textbf{Group 1} & \textbf{Group 2}& \textbf{Mean diff.} & \textbf{Lower diff.} & \textbf{Upper diff.}  \\ \hline 

0-13   &14–20  &  6.3042  & 6.2427   & 6.3657  \\   \hline

0-13 & 21-44 &-11.9353 & -11.9873 &-11.8833  \\  \hline 
 
0-13 &  45-90 & -19.8525 &  -19.9053&  -19.7997\\   \hline
 
14–20  & 21-44&  -18.2395  &  -18.2944 & -18.1845\\   \hline
 
14–20 &  45-90 & -26.1567  &  -26.2124 & -26.1009  \\ \hline

21-44  & 45-90   &-7.9172  &-7.9622  & -7.8722\\ \hline

\end{tabular}
\end{table}

\begin{figure}[!t]
\centering
    \includegraphics[scale=0.35]{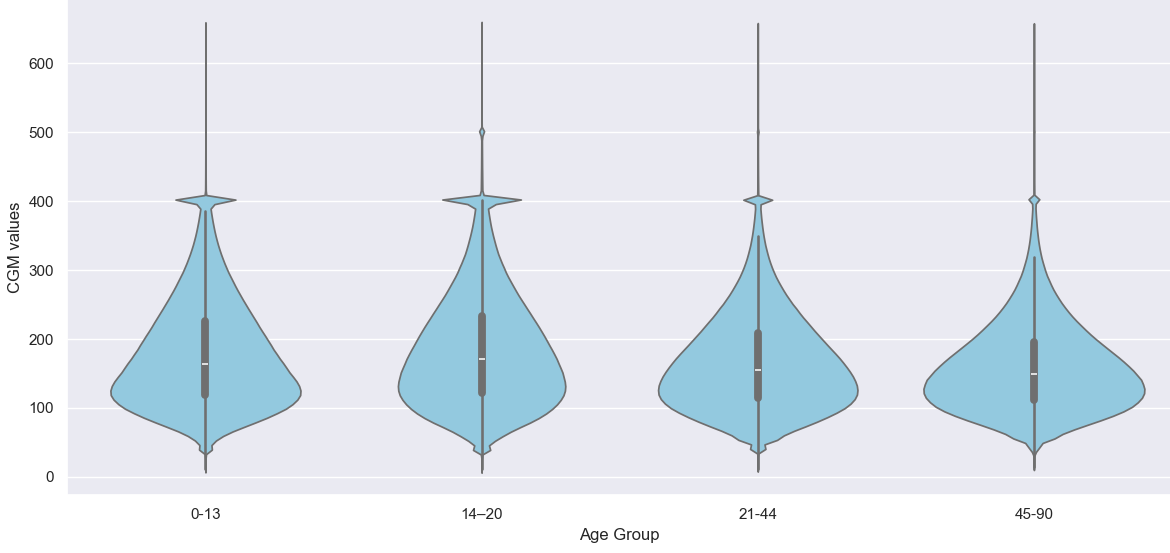}
\caption{Violin-plots per Age Group}
\label{fig:violinage}
\end{figure}

Fig. \ref{fig:violinage}, which presents violin plots per defined age group, shows that CGM variation differs across most groups in terms of the width of the graph, minimum, mean, and maximum values.
Finally, Table \ref{tab:Agestatistics} summarizes statistics of the age groups. Children and teenagers are underrepresented, with only half of the data of adults and seniors. 
\begin{table}[!ht]
\centering
\caption[Summary of Defined Age Groups]{Summary of Defined Age Groups \label{tab:Agestatistics}}
\begin{tabular}{c|c|c|c|c|}
\cline{2-5}
    & \textbf{0-13} & \textbf{14-20}& \textbf{21-44} & \textbf{45+}  \\ \hline 
    
\cline{2-5}
\multicolumn{1}{|l|}{\textbf{Subject count}} & 380 & 407 & 766 & 1035 \\ \hline

\cline{2-5}
\multicolumn{1}{|l|}{\textbf{Mean CGM}} & 178.89 & 185.19& 166.96 &   159.41
 \\   \hline

\cline{2-5}
\multicolumn{1}{|l|}{\textbf{STD of CGM}}&  78.17& 82.84& 70.22 & 64.60 \\   \hline 

\cline{2-5}
\multicolumn{1}{|l|}{\textbf{CGM value count}} & 19.8 mil &16.8 mil &  35.6 mil& 33.5 mil  \\ \hline

\end{tabular}
\end{table}

\subsection{Data Preprocessing}

After the classification of age groups, data were preprocessed for the classification task. 

\subsubsection{Data Cleaning and Imputation}

Fig. \ref{fig:violinage} reveals that the dataset comprises several CGM values out of the range of values in the confidence interval of CGM devices (40-500) \cite{Alva2023, Spartano2024}. Therefore, outliers were identified with the interquartile range (IQR) approach and replaced with missing values \cite{Torkey2021}, as well as values outside the range of 40-500 mg/dL. Missing values were grouped per gap duration based on the Impute Paradigm of Gupta et al., since CGM values do not behave linearly \cite{gupta_imavis, Gupta2025-Prag_Preprint, BIBE_vaibhav}. Then, short-term missing gaps ($\leq$25 min) were treated with linear interpolation and middle-term gaps (30-115 min) with the Stineman interpolation, which was also constrained to values in the range 40-500 mg/dL. Larger gaps were not imputed \cite{Cinar_BHI2025}.

\subsubsection{Class Assignment}

Following data cleaning, classes were assigned to classify hypoglycemia onset up to 120 minutes before. The framework for hypoglycemia classification is based on a two-step model, consisting of a binary classifier followed by a multi-class expert on hypoglycemic values, similar to D'Antoni et al. \cite{dantoni_prediction_2022}. The first model classifies between the risk of hypoglycemia 0-120 minutes before onset and between no risk in the preceding 120 minutes. The second model classifies the time range before the possible hypoglycemic event. This study foremost focuses on the multi-classification problem.

First, hypoglycemic data points are determined by a threshold below and equal to 70, and set as class 0. From each timestamp of class 0, the other time ranges are subtracted and assigned to the next class, ensuring no reassignments \cite{cinarMaster}. Previously, 5 classes were introduced, comprising 0, 5-10, 15-25, 30-55, and 60-120 minutes before event onset. Since earlier works reveal high misclassification rates, especially for classes 2 and 3 \cite{bibe_diadata_2025, Cinar_BHI2025}, this study also compared two new sets of classes. The first set includes 0, 5-15, 20-45, 50-120 before hypoglycemia, and the second set includes 0, 5-20, 25-60, 65-120 before event onset. Table \ref{tab:classesdistrtrain} presents the sample distribution of the classes for the train data. The no-risk class, defined as more than 120 minutes before hypoglycemia, was excluded from training.

\begin{table}[h!]
\centering
\caption{Distribution of Classes of the Train Set}
\label{tab:classesdistrtrain}
\begin{tabular}{|c|c|c|c|c|c|}
\hline
\textbf{Class} & \textbf{0}& \textbf{1}&\textbf{2}  & \textbf{3} & \textbf{4}    \\
\hline
\textbf{Set I} & 5.9 mil &1.2 mil&  1.7 mil& 3.1 mil&6.1 mil \\
\hline
\textbf{Set II} & 5.9 mil &  1.7 mil &3.2 mil &7.1 mil & N/A  \\ \hline
\textbf{Set III} &5.9 mil & 2.3 mil &4.1 mil& 5.6 mil  &  N/A \\ \hline
\end{tabular}
\end{table}

\begin{figure}[h]
\centering
    \includegraphics[scale=0.35]{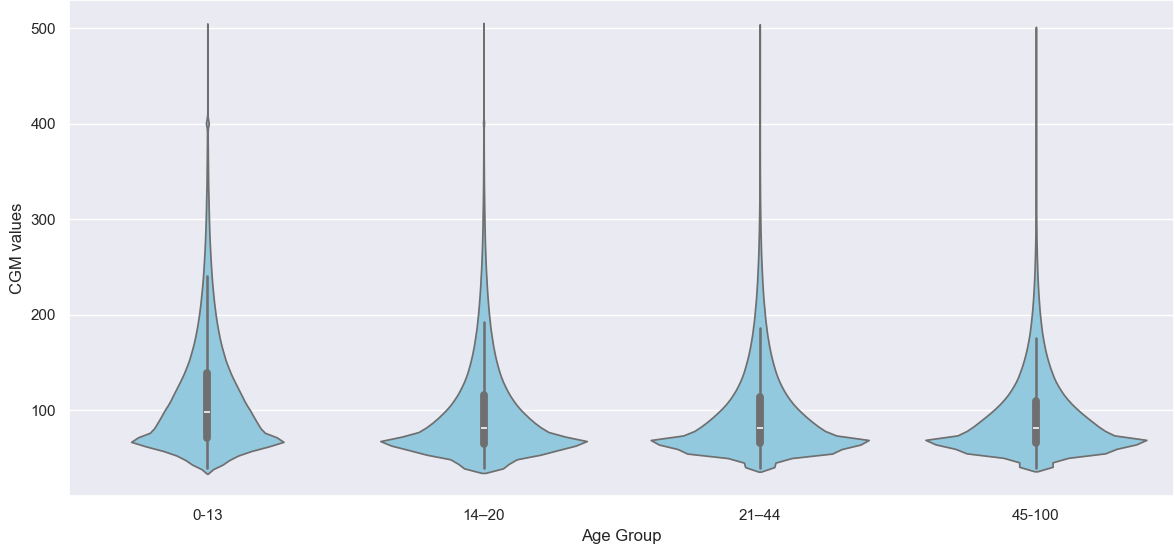}
\caption{Violin-plots per Age Group for Hypoglycemic Values}
\label{fig:violinagehypo}
\end{figure}

After data cleaning and class assignment, the violin plots of only hypoglycemic levels were explored, which are presented in Figure \ref{fig:violinagehypo}. The hypoglycemic datapoints are represented by all classes until 120 minutes before hypoglycemia. Figure \ref{fig:violinage} reveals that hypoglycemic patterns seem similar for teenagers, adults, and older adults. However, children have a narrower violin inducing that values are more tightly clustered.

\subsubsection{Time Series Generation and Data Splits}

To construct the test set, we selected the 10 subjects with the largest number of datapoints from each age group. In total, the test set comprised 40 subjects, whose characteristics are summarized in Table \ref{tab:test}, while the data of train subjects are displayed in Table \ref{tab:train}. No subject included in the test set appeared in the training set, even if associated with a different age group. After splitting subjects into train and test sets, CGM data were normalized between 0 and 1 with a global min-max scaler.

\begin{table}[h]
\centering
\caption[Summary of the Train Set]{Summary of the Train Set\label{tab:train}}
\begin{tabular}{|c|c|c|c|c|c|c|c|c|c|c|c|c|}
\hline 
    \textbf{Group} & \textbf{Mean d} & \textbf{Min d} & \textbf{Max d} & \textbf{Values}& \textbf{Mean Values}&
    \multicolumn{4}{ l |} {\textbf{Samples in class set II}}\\ 
    \cline{7-10} && & & && 0& 1  & 2 & 3  \\ \hline

\textbf{0-13} & 175 & 2& 361&18.7 mil  &  50681&610954 &286702 & 518626& 1.1 mil \\
\hline
\textbf{14-20}&  136 &0 &356&  15.6 mil &39418& 864349& 241667& 436475& 964866 \\\hline
\textbf{21-44} &  146  &0&1023&31.7 mil & 42075 &2.4 mil &634334 &1.2 mil &  2.6 mil\\\hline
\textbf{45+} & 109 & 1  &359& 32.4 mil & 31605& 2.1 mil& 579000&1.1 mil &  2.4 mil\\\hline
\end{tabular}
\end{table}

\begin{table}[h]
\centering
\caption[Summary of the Test Set]{Summary of the Test Set\label{tab:test}}
\begin{tabular}{|c|c|c|c|c|c|c|c|c|c|c|c|c|}
\hline 
    \textbf{Group} & \textbf{Mean d} & \textbf{Min d}& \textbf{Max d}& \textbf{Values}& \textbf{Mean Values}&
    \multicolumn{4}{ l |} {\textbf{Samples in class set II}}\\ 
    \cline{7-10} && &&& &  0& 1  & 2 & 3  \\ \hline

\textbf{0-13} & 376&364&412&1.1 mil&108543     & 18725&14235&26647& 60733\\
\hline
\textbf{14-20}&  390 &359 &455& 1.1 mil&112521&93489& 23566&42866&96496 \\ \hline
\textbf{21-44} &  1330 &1027& 1555 &3.8 mil&   383034&84207&35631& 64344& 145913   \\\hline
\textbf{45+} & 378 & 359&476&   1.1 mil &108942& 136125& 39177&71161&131870\\ \hline
\end{tabular}
\end{table}

For the ablation studies, a smaller subset was created by randomly selecting 10\% of the subjects in the train sets from each age group to reduce computational cost. Within this subset, 30\% of the subjects were assigned to the test set, and the remaining 70\% were used for training.

After splitting subjects into train and test sets, CGM data were normalized with a global min-max scaler to values between 0 and 1. Then, for each subject, time series were created with a sliding window approach. The input sequence length (ISL) was set to 30 minutes, corresponding to 7 data points \cite{Cinar_BHI2025}. Sequences with missing values and time gaps were removed. Finally, the sequence was labeled based on the class of the last instance in the sequence. 

Thereafter, the data of the test subjects were further split into train and test subsets using a 60:40 ratio for the fine-tuning approach. To minimize temporal dependency and introduce a temporal gap between train and test data, we removed the final 10\% of the train set's data samples. All splits were performed separately for each subject without shuffling. All model architectures were evaluated only on the test portion of the test subjects. 

Model individualization was tested since prior studies reported that population-based models later fine-tuned on the subject's data lead to improved performance \cite{Neumann2025-te}.

\subsection{Model Architecture}

The model was trained with the Fully Convolutional Network (FCN) model of Wang et al. \cite{wang_time_2016} consisting of 3 convolutional layers. The size of 1DCNN layers was 128, 256, and 128 with kernel sizes of had kernel sizes 8, 5, and 3, respectively. All 1DCNN layers were followed by batch normalization and a ReLU activation layer \cite{wang_time_2016}.
The last convolution was also followed by a global pooling layer and the activation layer.

\subsection{Ablation Studies}

To tune the model, ablation studies were conducted. As an initial starting point, we employed our previously optimized architecture, which uses a 30-minute ISL with class set I and uses focal loss to mitigate class imbalance \cite{Cinar_BHI2025}. The models were trained with a batch size of 128 for 20 epochs over three seed numbers (48, 0, 1234): 1) We compared a default focal loss with a focal loss with adjusted class weights for class set I based on a balanced class distribution. 2) Then, we compared different class sets (I, II, and III) with the better loss function. 3) The better class set was further tested with various ISLs of 45, 60, 90, and 120 minutes, corresponding to 10, 13, 19, and 25 consecutive datapoints, respectively. 4) In addition, the better set of classes was trained with a 15-minute sampling rate with ISLs of 30, 45, 90, 60, and 120 minutes. Data were first rounded to 5 and then oversampled to 15 minutes for each subject.

\section{Results}
\label{sec:results}

After the ablation studies, the best-performing model was trained on all subjects in the training set for the global population-based (GPB) model using a batch size of 512. Then, age-specific population-based (ASPB) models were trained on all training subjects within each age group using a batch size of 264. All models were trained for 100 epochs \cite{ghimire_deep_2024} and evaluated on the last 40\% of each test subject's data. The models were further fine-tuned for 5 epochs using the training portion of each test subject and evaluated on the same test portion.

\subsection{Ablation Studies}

The results of the ablation studies are summarized in Table \ref{tab:AblationStudies}. Training with focal loss (FL) using class-specific alpha weights (w) improves recall compared to the model trained with the default settings. 
Using the default focal loss, the precision is increased at the cost of lower recall. Clinically, misclassifying classes 3 or 4 is less critical, whereas correctly identifying events occurring shortly before hypoglycemia is essential. Consequently, adjusted alpha weights were incorporated in the focal loss.

Furthermore, the model can better distinguish the bins of class set II (0, 5-15, 20-45, 50-120 min before onset). Among all configurations, an ISL of 120 minutes yields the best performance for recall, F1-M, and PR-AUC. Finally, a 5-minute sampling rate improves performance by up to 5\% relative to a 15-minute sampling rate. For the 15-minute sampling rate, multiple ISLs (I:30, II:45, III:60, IV:90, V:120) were evaluated, with the best performance achieved using a 120-minute window.

\begin{table}[h!]
\centering
\caption[Macro Results of the Ablation Studies]{Macro Results of the Ablation Studies\label{tab:AblationStudies}}
\begin{tabular}{c|c|c|c|c|}
\cline{2-5}
     & \textbf{Recall}& \textbf{Precision}& \textbf{F1-M} & \textbf{PR-AUC}  \\ \hline

 \cline{2-5}
\multicolumn{1}{|l|}{\textbf{FL default}} & 0.7051 & 0.7039 &   0.7005 &  0.7398
 \\   
 \hline \hline

\cline{2-5}
\multicolumn{1}{|l|}{\textbf{Class I + FL + w}}  & 0.7288 & 
0.6826  & 0.6984 & 0.7319 
 \\   
 \hline 

\cline{2-5}
 \multicolumn{1}{|l|}{\textbf{Class II + FL + w}}   &  \textbf{0.8019} & \textbf{0.7598} &  \textbf{0.7727}&\textbf{0.8185}\\   \hline 

\cline{2-5}
\multicolumn{1}{|l|}{\textbf{Class III + FL + w}} & 0.7747 & 0.7518 &  0.7592 & 0.8019 \\   \hline \hline

\cline{2-5}
\multicolumn{1}{|l|}{\textbf{ISL of 45 min}} &
0.8048& 0.7656 &   0.7782&0.8260\\   \hline

\cline{2-5}
 \multicolumn{1}{|l|}{\textbf{ISL of 60 min}} &  0.8053 &
0.7642  & 0.7770&0.8247  \\   \hline

\cline{2-5}
 \multicolumn{1}{|l|}{\textbf{ISL of 90 min}} &0.8047 & 
\textbf{0.7728}  &  0.7829& 0.8273 \\   \hline

\cline{2-5}
\multicolumn{1}{|l|}{\textbf{ISL of 120 min}}  &\textbf{0.8074}  &
0.7716   & \textbf{0.7844}  &\textbf{0.8290}\\    \hline  \hline

\cline{2-5}
\multicolumn{1}{|l|}{\textbf{15 min sampling I}}  &0.7522 &  0.7133  & 0.7225  &  0.7580    \\  \hline

\cline{2-5}
\multicolumn{1}{|l|}{\textbf{15 min sampling II}}  &  0.7548  &   0.7138 &   0.7254   &0.7673   \\  \hline

\cline{2-5}
\multicolumn{1}{|l|}{\textbf{15 min sampling III}}  &  0.7679  & 0.7286  &   0.7401   &  0.7754 \\  \hline

\cline{2-5}
\multicolumn{1}{|l|}{\textbf{15 min sampling IV}}  &  0.7720  &0.7283&  0.7384    & 0.7744  \\  \hline

\cline{2-5}
\multicolumn{1}{|l|}{\textbf{15 min sampling V}}  & \textbf{0.7727} &  \textbf{0.7308} &   \textbf{0.7439}  & \textbf{0.7761}   \\  \hline

\end{tabular}
\end{table}

The final distribution for class set II of the test dataset is presented in Table \ref{tab:classesdistr}.
\begin{table}[h!]
\centering
\caption{Distribution of Classes of the Test Set}
\label{tab:classesdistr}
\begin{tabular}{|c|c|c|c|c|}
\hline
\textbf{Class} & \textbf{0}& \textbf{1}&\textbf{2}  & \textbf{3}      \\
\hline

\textbf{Set II} &332546 &  112609 &205018 &435012   \\ \hline

\end{tabular}
\end{table}

\subsection{Individualization vs. Population-Based Approaches}

\begin{table}[!ht]
\centering
\caption[Performance of FCN Models]{Performance of FCN Models\label{tab:Comp5classles}}
\begin{tabular}{c|c|c|c|c|c|c|c|}
\cline{2-7}
    & \textbf{Metric} & \textbf{M-avg} &
    \multicolumn{4}{ l |} {\textbf{Class}}\\ 
    \cline{4-7} & & & 0& 1  & 2 & 3  \\ \hline

\multicolumn{1}{|l|}{\multirow{6}{*}{}} &   Recall & 0.7903   & \textbf{1.0000}   & 0.7927   & 0.5916  &  \textbf{0.7770}  \\

\cline{2-7}  
\multicolumn{1}{|l|}{} & Precision &  \textbf{0.7720}   &  \textbf{0.9997}   & \textbf{0.7093}  & \textbf{0.5272}   & 0.8519  \\

\cline{2-7}        
\multicolumn{1}{|l|}{\textbf{GPB}}   & F1-M  & \textbf{0.7785} & \textbf{0.9998}   &\textbf{0.7469} &0.5558  &\textbf{0.8114}   \\

\cline{2-7}    
\multicolumn{1}{|l|}{}   & PR-AUC      & \textbf{0.8159}  &  \textbf{1.0000} &   \textbf{0.8156} & \textbf{0.5451} & \textbf{0.9028}   \\
\hline \hline

\multicolumn{1}{|l|}{\multirow{6}{*}{}}  &  Recall &   0.7892  &0.9953   & 0.8048   &\textbf{0.6172}  &0.7394 \\
\cline{2-7}   
\multicolumn{1}{|l|}{} & Precision   & 
0.7627   &0.9975 & 0.6804  &0.5082  & \textbf{0.8646}     \\

\cline{2-7}   
\multicolumn{1}{|l|}{\textbf{GPB + I}} & F1-M  &  0.7707  &0.9963  & 0.7350  & \textbf{0.5560}  & 0.7957   \\

\cline{2-7}    
\multicolumn{1}{|l|}{}  & PR-AUC   &    0.8073   & \textbf{1.0000}  &0.7995 & 0.5292  &0.9006   \\ \hline \hline

\multicolumn{1}{|l|}{\multirow{6}{*}{}} &  Recall & \textbf{0.7915} & \textbf{1.0000}  & \textbf{0.8337}   & 0.5879  &0.7443    \\ 

\cline{2-7}   
\multicolumn{1}{|l|}{} & Precision& 0.7580 & 0.9981  &0.6562  &0.5134  &0.8643  \\

\cline{2-7}     
\multicolumn{1}{|l|}{\textbf{ASPB}}   & F1-M  &  0.7687   & 0.9990  & 0.7317   & 0.5463  & 0.7978 \\

\cline{2-7}    
\multicolumn{1}{|l|}{}  & PR-AUC &0.8099  &\textbf{1.0000}   & 0.8071 & 0.5302 & 0.9023   \\ \hline \hline

\multicolumn{1}{|l|}{\multirow{6}{*}{}} & Recall &  0.7849 & 0.9969  & 0.8027 & 0.6058 & 0.7341  \\

\cline{2-7}     
\multicolumn{1}{|l|}{}  & Precision  &  0.7593  & 0.9983  & 0.6787  &0.5004 & 0.8598    \\

\cline{2-7}  
\multicolumn{1}{|l|}{\textbf{ASPB + I}} & F1-M  & 0.7667  & 0.9975  & 0.7328  &0.5461  &  0.7901   \\

\cline{2-7}  
\multicolumn{1}{|l|}{}  & PR-AUC  &  0.8016  & \textbf{1.0000}     & 0.7921 &0.5170  & 0.8973    \\
\hline

\end{tabular}
\end{table}

\begin{table}[!ht]
\centering
\caption[Performance per Age Group]{Performance per Age Group\label{tab:AgeBased}}
\begin{tabular}{c| c|c|c|c|c|}
\cline{2-6}
    &\textbf{Age Group}& \textbf{Recall} & \textbf{Precision}& \textbf{F1-M} & \textbf{PR-AUC}  \\ \hline

 \cline{2-6}  
\multicolumn{1}{|l|}{} & \multicolumn{1}{|l|}{Children} &   0.7666 & \textbf{0.7544}  & \textbf{0.7592} & \textbf{0.7941} 
 \\

 \cline{2-6}  
\multicolumn{1}{|l|}{} &\multicolumn{1}{|l|}{Teenagers} & \textbf{0.8279}  & \textbf{0.8012} &  \textbf{0.8106}  &  \textbf{0.8535}  
 \\

 \cline{2-6}  
\multicolumn{1}{|l|}{\textbf{GPB}} &\multicolumn{1}{|l|}{Adults}& \textbf{0.7768}   &  \textbf{0.7586} &  \textbf{0.7654}  &  \textbf{0.7986}  
 \\

 \cline{2-6}  
\multicolumn{1}{|l|}{} &\multicolumn{1}{|l|}{Seniors} & \textbf{0.7901}  & \textbf{0.7739} & \textbf{0.7787} &\textbf{0.8173} \\   
 \hline  \hline

\cline{2-6}
\multicolumn{1}{|l|}{} &\multicolumn{1}{|l|}{Children}  & \textbf{0.7845}  & 0.7389   & 0.7520  & 0.7934  
 \\

\cline{2-6}
\multicolumn{1}{|l|}{} &\multicolumn{1}{|l|}{Teenagers} & 0.8201  &  0.7843   & 0.7978  & 0.8446  
 \\

\cline{2-6}
\multicolumn{1}{|l|}{\textbf{ASPB}} &\multicolumn{1}{|l|}{Adults} & \textbf{0.7768}  & 0.7432  & 0.7530   & 0.7881  
 \\   
 
\cline{2-6}
\multicolumn{1}{|l|}{} &\multicolumn{1}{|l|}{Seniors} &0.7844  & 0.7656  &  0.7721 &  0.8134 \\  \hline 

\end{tabular}
\end{table}
Comparing the performance of the global population-based (GPB) and the age-segmented population-based (ASPB) models, which are depicted in Table \ref{tab:Comp5classles}, it can be seen that the former yields better results across all metrics except the recall of class 1. Nevertheless, in most cases, the GPB model is only minimally superior, showing improvements of 0.60-1.40\% in the macro-averaged metrics. This indicates that even with less data, the model performs almost as well as a GPB model, if specialized on age groups. 

However, fine-tuning the model for individualization (I) does not provide a significant advantage and decreases most of the metrics for both the global and age-segmented population models. 

Table \ref{tab:AgeBased} presents the results of the global population-based and age-segmented models tested separately for each age group. Likewise, the GPB model performs similarly or better in most metrics. The macro-averaged recall of teenagers, adults, and seniors is increased, and the macro-averaged precision of children, teenagers, and seniors is notably increased. In contrast, the recall of children is increased by 1.79\% in the ASPB model. For the F1-M and PR-AUC, all groups have better values with the GPB model. Consequently, age-segmented models demonstrate superior recall for data of children.
  
\begin{figure}[h]
\centering

\begin{subfigure}{0.26\textwidth}
    \includegraphics[width=\textwidth]{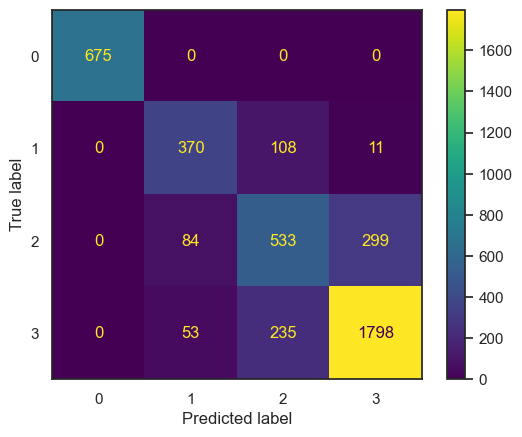}
    \caption{Best Result Children}
     
\end{subfigure}
\begin{subfigure}{0.26\textwidth}
    \includegraphics[width=\textwidth]{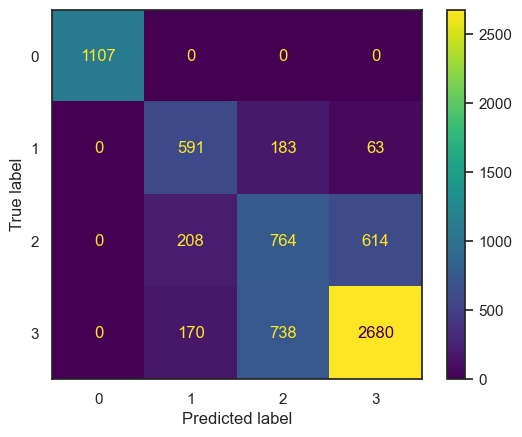}
    \caption{Worst Result Children}
     
\end{subfigure}

\begin{subfigure}{0.26\textwidth}
    \includegraphics[width=\textwidth]{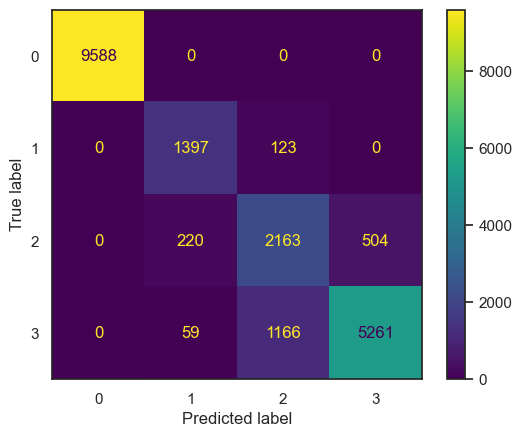}
    \caption{Best Result Teenagers}
      
\end{subfigure}
\begin{subfigure}{0.26\textwidth}
    \includegraphics[width=\textwidth]{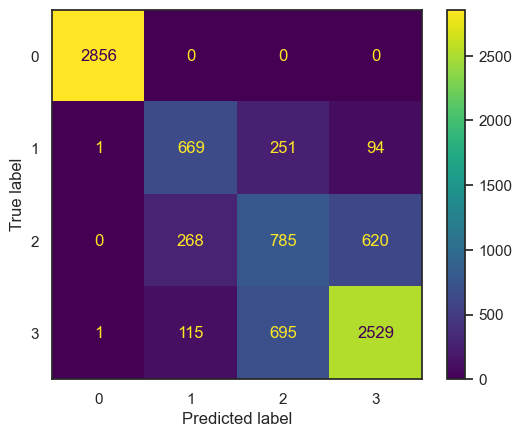}
    \caption{Worst Result Teenagers}
     
\end{subfigure}

\begin{subfigure}{0.26\textwidth}
    \includegraphics[width=\textwidth]{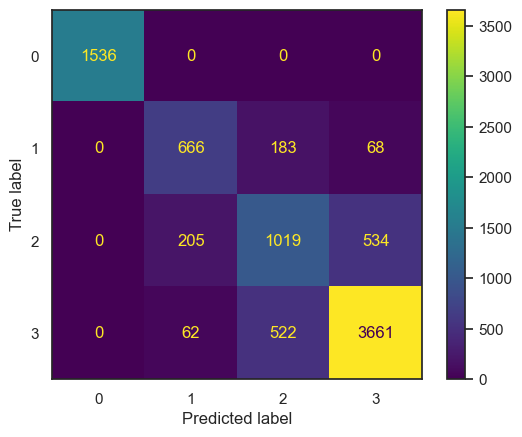}
    \caption{Best Result Adults}
      
\end{subfigure}
\begin{subfigure}{0.26\textwidth}
    \includegraphics[width=\textwidth]{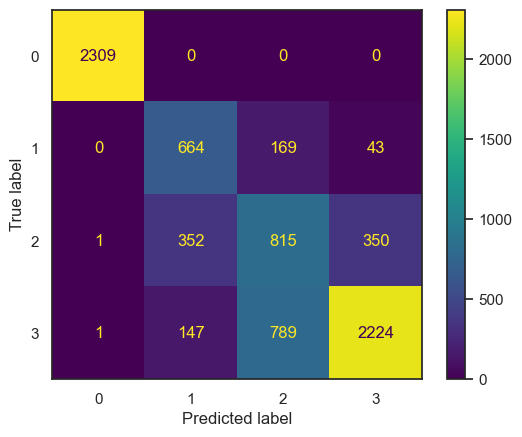}
    \caption{Worst Result Adults}
     
\end{subfigure}

\begin{subfigure}{0.26\textwidth}
    \includegraphics[width=\textwidth]{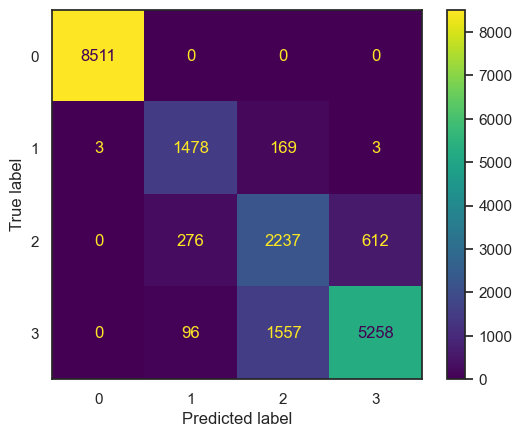}
    \caption{Best Result Seniors}
      
\end{subfigure}
\begin{subfigure}{0.26\textwidth}
    \includegraphics[width=\textwidth]{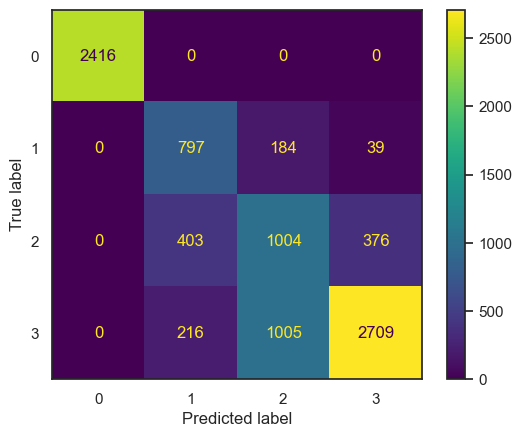}
    \caption{Worst Result Seniors}
     
\end{subfigure}

\caption{Confusion Matrices}
\label{fig:CM}
\end{figure}

Fig. \ref{fig:CM} reveals the confusion matrices of the subjects with the best and worst performance trained with the GPB model for each age group, while Table \ref{tab:CMresAG} reveals their corresponding metrics. The best macro scores are obtained for the second age group (teenagers). The worst scores are also for the second age group. In both cases, the recall of classes 0, 1, and 3 is increased. The precision is decreased in the worst test fold, and the model fails to distinguish between classes 2 and 3. The other confusion matrices show similar patterns, especially for the best performances. In all cases, the precision is less than the recall. For the worst performances, the model has difficulties in differentiating between classes 2 and 3 for instances belonging to class 3.

\begin{table}[h]
\centering
\caption[Best and Worst Performances per Age Group]{Best and Worst Performances per Age Group\label{tab:CMresAG}}
\begin{tabular}{c| c|c|c|c|c|}
\cline{2-6}
    &\textbf{Age Group}& \textbf{Recall} & \textbf{Precision}& \textbf{F1-M} & \textbf{PR-AUC}  \\ \hline

 \cline{2-6}  
\multicolumn{1}{|l|}{} & \multicolumn{1}{|l|}{Children}   & 0.8001  &   0.7978 &  0.7988   &   0.8537
 \\

 \cline{2-6}  
\multicolumn{1}{|l|}{} &\multicolumn{1}{|l|}{Teenagers} &  0.8699   &   0.8432   & 0.8539   & 0.9039
 \\

 \cline{2-6}  
\multicolumn{1}{|l|}{\textbf{Best Performance}} &\multicolumn{1}{|l|}{Adults}&  0.7921 & 0.7909    &  0.7915 & 0.8264
 \\

 \cline{2-6}  
\multicolumn{1}{|l|}{} &\multicolumn{1}{|l|}{Seniors} &  0.8427           & 0.8146   &  0.8244 &0.8804\\   
 \hline  \hline

\cline{2-6}
\multicolumn{1}{|l|}{} &\multicolumn{1}{|l|}{Children}  &   0.7337  &    0.7154    &     0.7233   & 0.7474
 \\

\cline{2-6}
\multicolumn{1}{|l|}{} &\multicolumn{1}{|l|}{Teenagers} &  0.7214  &  0.7171   &     0.7191     & 0.7452
 \\

\cline{2-6}
\multicolumn{1}{|l|}{\textbf{Worst Performance}} &\multicolumn{1}{|l|}{Adults} &  0.7496 & 0.7199     &     0.7290    &  0.7600  
 \\   
 
\cline{2-6}
\multicolumn{1}{|l|}{} &\multicolumn{1}{|l|}{Seniors} & 0.7584 &  0.7220    &    0.7319  &  0.7646\\  \hline 

\end{tabular}
\end{table}

\section{Discussion}
\label{sec:discussion}

To the knowledge of the authors, this is the first study investigating the impact and generalizability of age-segmented models and population-based models for hypoglycemia classification leveraging DiaData, comprising all age categories of patients with T1D. Previous studies only investigated the impact of age on glucose forecasting.  

First, based on graphical and statistical data analysis, we classified subjects into age groups of children (0-13), teenagers (14-20), adults (20-44), and seniors (45-100). Exploratory visualization of box-, violin-, and lineplots, and Tukey's analysis revealed significant differences in CGM values and patterns across the age groups. Furthermore, our previous work demonstrated that the age group is the most predictive feature among personal values for hypoglycemia classification \cite{bibe_diadata_2025}. These results suggest that separate age groups could require expert models, as also advised by related studies \cite{So2022-ze}.

Training the models, our results reveal that a global population-based model can generalize effectively across age groups, and training with a large, heterogeneous dataset leads to superior results. This indicates that hypoglycemia reveals similar patterns and possibly similar causes across age groups, which is why the model learns better with increased data volume and a variety of subjects. Furthermore, age-segmented population-based models also yielded comparable performance with minimally decreased values. Thus, even if less data is available, the models specialized on age groups can classify hypoglycemia almost as well as global population-based models. Testing model individualization, the results decreased. Even when selecting subjects with high data volume, the model may still have been influenced by temporal dependencies in the training data, leading to overfitting to temporal patterns. 

Since our analysis revealed differences in CGM variation, future work should explore age-segmented models with glucose forecasting. Moreover, various deep learning and machine learning models should be compared to increase model performance. In addition, to mitigate data imbalance, proper data quality should be addressed \cite{gupta_imavis}. An optimized model could be integrated into an app to assist patients in hypoglycemia prevention \cite{Grensing2025EarlyWarning}.

\section{Conclusion}
\label{sec:conclusion}

This work explored the impact and generalizability of age-segmented and global population-based models for hypoglycemia classification. We defined age groups of 1-13, 14-20, 21-44, and above 44 years based on statistical data analysis. All data clusters, including the global population using all data, and the age-specific subsets, were trained with the same FCN model. The models were tested on the same hold-out set of 10 unseen subjects from each age group. The results reveal that global population-based models yield better performance, while age-segmented models show similar classification ability. However, individualization via transfer learning decreases the performance. Consequently, if more data are available, population-based models can be used to increase stability, even if some age ranges are underrepresented. A model specialized in less data of a particular age group performs almost as well. For maximal recall in children's data, age-specialized models are suggested.


%%%%%%%%%%%%%%%%%%%%%%%%%%%%%%%%%%%%%%%%%%%%%%%%%%%%%%%%%%%%%%%%%%%%%%%%%%%%%%%%

%%%%%%%%%%%%%%%%%%%%%%%%%%%%%%%%%%%%%%%%%%%%%%%%%%%%%%%%%%%%%%%%%%%%%%%%%%%%%%%%

%%%%%%%%%%%%%%%%%%%%%%%%%%%%%%%%%%%%%%%%%%%%%%%%%%%%%%%%%%%%%%%%%%%%%%%%%%%%%%%%

%%%%%%%%%%%%%%%%%%%%%%%%%%%%%%%%%%%%%%%%%%%%%%%%%%%%%%%%%%%%%%%%%%%%%%%%%%%%%%%%

\section*{Data Statement}

The sources of subsets of the data are the Barbara Davis Center, Jaeb Center for Health Research, Joslin Diabetes Center, T1D Exchange, University of Colorado, and the University of Virginia. Retrieved from: https://public.jaeb.org/dataset. The analyses, content, and conclusions presented herein are solely the responsibility of the authors and have not been reviewed or approved by the aforementioned institutions. A part of the dataset can be downloaded from https://zenodo.org/records/16874128~\cite{DiaData_Zenodo}.

\bibliographystyle{ieeetr}
\bibliography{references}

\end{document}